\documentclass[sigconf,nonacm]{acmart}

\setcopyright{none}
\renewcommand\footnotetextcopyrightpermission[1]{}
\AtBeginDocument{%
  }

\usepackage{hyperref}
\usepackage{url}
\usepackage{algorithmic}
\usepackage{booktabs}
\usepackage{booktabs}       % professional-quality tables
\usepackage{amsfonts}       % blackboard math symbols
\usepackage{nicefrac}       % compact symbols for 1/2, etc.
\usepackage{microtype}      % microtypography
\usepackage{xcolor}         % colors
\usepackage{amsmath} 
\usepackage{graphicx} 
\usepackage{multirow}
\usepackage{tabularx}
\usepackage{xspace}
\usepackage{tabularx}
\usepackage[most]{tcolorbox}
\usepackage[ruled,vlined]{algorithm2e}

\newcommand{\tool}{X-RAY\xspace}

\copyrightyear{2026}
\acmYear{2026}
\setcopyright{cc}
\setcctype{by}
\acmConference[KDD '26]{Proceedings of the 32nd ACM SIGKDD Conference on Knowledge Discovery and Data Mining V.2}{August 09--13, 2026}{Jeju Island, Republic of Korea}
\acmBooktitle{Proceedings of the 32nd ACM SIGKDD Conference on Knowledge Discovery and Data Mining V.2 (KDD '26), August 09--13, 2026, Jeju Island, Republic of Korea}
\acmDOI{10.1145/3770855.3818029}
\acmISBN{979-8-4007-2259-2/2026/08}

\begin{document}

% \title{Mapping LLM Capability Frontiers: A Data-Driven Exploration with Calibrated and Verified Probes}
% \title{Quantifying LLM Capability Frontiers via Calibrated and Verified Probes}
\title{Mapping LLM Capability Frontiers via Formalized and Calibrated Probes}
\settopmatter{authorsperrow=4}

\author{Tianxi Gao}
\email{gaotianxi@u.nus.edu}
\affiliation{%
  \institution{National University of Singapore}
  \country{Singapore}
}

\author{Yufan Cai}
\email{cai_yufan@u.nus.edu}
\affiliation{%
  \institution{National University of Singapore}
  \country{Singapore}
}
% \authornotemark
\authornote{Corresponding author}

\author{Yusi Yuan}
\email{yusiyuan@u.nus.edu}
\affiliation{%
  \institution{National University of Singapore}
  \country{Singapore}
}

\author{Jin Song Dong}
\email{dcsdjs@nus.edu.sg}
\affiliation{%
  \institution{National University of Singapore}
  \country{Singapore}
}

\begin{abstract}
Large language models (LLMs) achieve promising performance, yet their ability to reason remains poorly understood.
Existing evaluations largely emphasize task-level accuracy, often conflating pattern matching with reasoning capability.
We present X-RAY, an \textit{eXplainable Reasoning Analysis sYstem} that maps the LLM reasoning capability using calibrated, formally verified probes.
We model reasoning capability as a function of extractable \textit{structure}, operationalized through formal properties such as constraint interaction, reasoning depth, and solution-space geometry.
X-Ray generates probes via formal tools with controlled structural variations, enabling isolation of structural information through calibration and verification.
We evaluate state-of-the-art LLMs on problems ranging from junior-level to advanced in mathematics, physics, and chemistry.
Our analysis reveals a systematic asymmetry in LLM reasoning: models are relatively robust to constraint refinement, in which additional conditions shrink an existing solution space, but degrade sharply under solution-space restructuring, in which modifications alter the underlying structural form of the solution manifold. 
Moreover, calibrated formal probes distinguish models that appear indistinguishable on standard benchmarks and reveal interpretable rather than opaque failure modes.
Beyond evaluation, our framework is contamination-resistant and supports the training and testing of reasoning models.
\end{abstract}




\maketitle

% \newcommand\kddavailabilityurl{https://doi.org/10.34740/kaggle/dsv/16590165}
% \ifdefempty{\kddavailabilityurl}{}{
% \begingroup\small\noindent\raggedright\textbf{Resource Availability:}\\
% % please change the following context to include multiple artifacts if necessary, including data, models, code, etc.
% The artifact of this paper has been made publicly available at \url{\kddavailabilityurl}.
% \endgroup
% }

\section{Introduction}
\label{sec:intro}
Large language models (LLMs) have demonstrated impressive performance on a wide range of reasoning benchmarks, spanning arithmetic, symbolic manipulation, and multi-step problem solving~\cite{zhong2024gsm8k,wei2023chain}.
Yet these results leave a fundamental question unresolved: \emph{what reasoning capacity do these models actually possess, and where are their limits?}
Most existing evaluations~\cite{livebench,hendrycks2021math} report task-level accuracy on fixed datasets, offering limited insight into how models generalize beyond seen instances or why performance degrades under more demanding conditions.
As a result, benchmark scores often conflate structured reasoning ability with pattern matching, serving more as leaderboards than as instruments for measuring reasoning capacity.

\paragraph{From pattern matching to structured reasoning.}
High accuracy on a reasoning benchmark does not necessarily imply structured reasoning capability~\cite{dziri2023faith,razeghi2022impactpretrainingtermfrequencies}.
When evaluations primarily vary surface form, such as lexical diversity or problem phrasing, 
models may succeed by matching familiar templates rather than by extracting and recomposing latent constraints.
In contrast, structured reasoning requires robustness to novel combinations of conditions, dependencies, and reasoning paths, precisely where pattern-based generalization tends to break down.
From this perspective, task difficulty is not determined by raw entropy or input length, but by the amount of \emph{structure} that must be extracted by a computationally bounded learner, echoing recent views on structural information~\cite{finzi2026entropyepiplexityrethinkinginformation}.
This distinction becomes critical in problems involving conditional constraints or multi-step transformations, where pattern matching provides little guidance.

\paragraph{Why formal verification.}
Moving from empirical evaluation to principled capability measurement requires reasoning probes with unambiguous semantics and reliable ground truth.
However, many existing benchmarks suffer from annotation noise, latent ambiguities, and uncontrolled surface cues, which can dominate measured accuracy~\cite{liang2023holisticevaluationlanguagemodels}.
These issues are further compounded by dataset contamination~\cite{carlini2023quantifying}, making improvements on static benchmarks increasingly difficult to interpret.
Consequently, probing reasoning demands benchmarks that are
(i) calibrated to eliminate surface-level confounders, and
(ii) formally verified to guarantee correctness. 
Verifiers~\cite{de2008z3,barbosa2022cvc5,mathematica14} provide a natural foundation for such probes by ensuring semantic well-posedness. 
It also enables cross-validation across different formal methods for a given problem.

\paragraph{Reasoning capability of LLMs.}
% 如何去形式化地定义这个边界？
We conceptualize reasoning ability not as a single scalar score but as a capacity exercised across increasingly complex structural requirements. 
From this perspective, a central question is how model performance evolves as structural complexity grows: 
does performance degrade gradually, or does it exhibit qualitative shifts under stronger structural coupling?
We treat reasoning capability as a structurally conditioned phenomenon.
We therefore analyze performance as a function of explicitly parameterized structural dimensions, 
such as constraint interaction depth or solution-space transformation. 
This perspective allows us to move beyond aggregate accuracy and localize failure modes to specific classes of structural operations.
To make this notion operational, task structure must be parameterizable and controllable. 
By constructing probes whose difficulty increases along explicit structural dimensions—and formally verifying their correctness—we create a controlled environment in which observed performance changes can be attributed to reasoning capability.

\paragraph{Our approach.}
We present \tool, an \emph{eXplainable Reasoning Analysis sYstem} for mapping LLMs' reasoning capability using formalized and calibrated probes.
Each probe is generated via structural transformations, 
like deepening compositional structure or strengthening cross-constraint coupling,
while formally preserving correctness and enabling automatic verification.
This design isolates the incremental \emph{structural information} required to solve a probe.
It also renders failures structurally interpretable: when a model breaks, failure can be attributed to specific structural factors rather than opaque dataset artifacts.
We apply our probes across diverse domains, including mathematics, physics, and chemistry, spanning junior- to advanced-level variants.
Finally, by parameterizing tasks along explicit structural dimensions, \tool naturally supports fine-tuning of reasoning models, enabling curricula that progressively expand extractable structure and diagnose brittle reasoning operations.
%%%%%%%%%%%%%
Across multiple LLM families, we observe clear and reproducible differences in capabilities under structurally distinct transformations.
In particular, reasoning LLMs like \textsc{o4-mini} tend to remain relatively stable under \emph{constraint refinement}, 
where additional conditions restrict an existing solution space without altering its underlying representation.
In contrast, performance degrades more substantially under \emph{solution-space restructuring}, 
where modifications require changes to the geometry or representation of the solution manifold itself.
This asymmetry indicates that reasoning performance is sensitive not merely to problem size or surface difficulty, 
but to the nature of the structural operation involved.
Because our probes are generated through programmatic transformations and formally verified by solvers, the resulting evaluation minimizes contamination and ensures that observed performance differences are attributable to structural variation.

\paragraph{Contributions.}
We make the following contributions:
\begin{itemize}
    \item \textbf{Structured reasoning capability.}
    We reformulate LLM evaluation as a problem of measuring how much structural information a model can extract and manipulate.
    
    \item \textbf{Formally calibrated probe construction.}
    We propose a probe construction pipeline, which preserves latent structure while removing superficial cues, with correctness guaranteed by formal methods.

    \item \textbf{Capability frontier discovery.}
    We show that LLM performance undergoes sharp phase-transition-like changes along controlled structural dimensions, producing reproducible capability frontiers that reveal reasoning limits and failure regimes.
    
    \item \textbf{A reusable evaluation and training substrate.}
    The proposed framework is contamination-resistant by construction and supports dynamic evaluation and downstream reasoning model training and testing.
\end{itemize}
 % 1.5
\section{Motivating Examples}
% Difficulty Calibration：不是 X1 放大，而是 program 组合
% Formal verification 是工具，不是目的：solver 只负责“裁判”
% online evaluation
% 最后一个例子能提供额外每个步骤的验证条件，展示拓展性，放进附录中！！！

To illustrate the core idea of our framework, we present several representative examples.

\subsection{N-Primable Numbers}
Consider the following simplified question:
\emph{``A positive number is called $n$-primable if it is divisible by $n$ and each of its digits is a one-digit prime number. How many 3-primable positive integers are there that are less than 1000?"}

\paragraph{Autoformalization.}
The autoformalizer maps this natural language into a formal constraint model. 
The parameterizer transforms it into a more general problem and builds the binds between formal code and natural language as:
The LLM encodes this with Z3\cite{de2008z3} as:
\begin{verbatim}
s = Solver()
X1, X2 = 1000, 3
s.add(number < X1)
s.add(number % X2 == 0)
...
\end{verbatim}
\tool validates the generated formalization by checking consistency across multiple formal methods and formal reasoning models.

\paragraph{Structural Calibration.}
\tool increases the compositional density of the problem by semantic constraints while keeping the formal skeleton unchanged.
For example, \textit{(1) One Extra Condition}: The sum of digits $\sum d_i$ must itself be a prime number, and \textit{(2) Two Extra Conditions}: An additional structural ordering is imposed such that $d_2 \le d_1 \le d_0$.
This calibration controls the logical coupling, and the variants require the model to simultaneously synchronize divisibility, set membership, and positional dependencies.

\paragraph{Probing.}
% The experimental procedure is similar to that of the first motivating example.
\tool randomizes $(X_1,X_2)$ within bounded domains to obtain a batch of solver-verified instances sharing the same latent structure.
For each randomization, we evaluate the problem with a targeted large language model 10 times.
Here, \emph{calibration} provides meaningful, localized difficulty adjustment (small structural shifts without changing the formal program skeleton). 
In contrast, \emph{randomization} provides the number of instances required for quantitative capability mapping (e.g., success-rate surfaces and phase-transition boundaries).
Because every generated instance is solver-checked, observed performance changes can be attributed to controlled variation in $(X_1,X_2)$ and the structural variance.

\begin{figure}[t]
    \centering
    \includegraphics[width=1\linewidth]{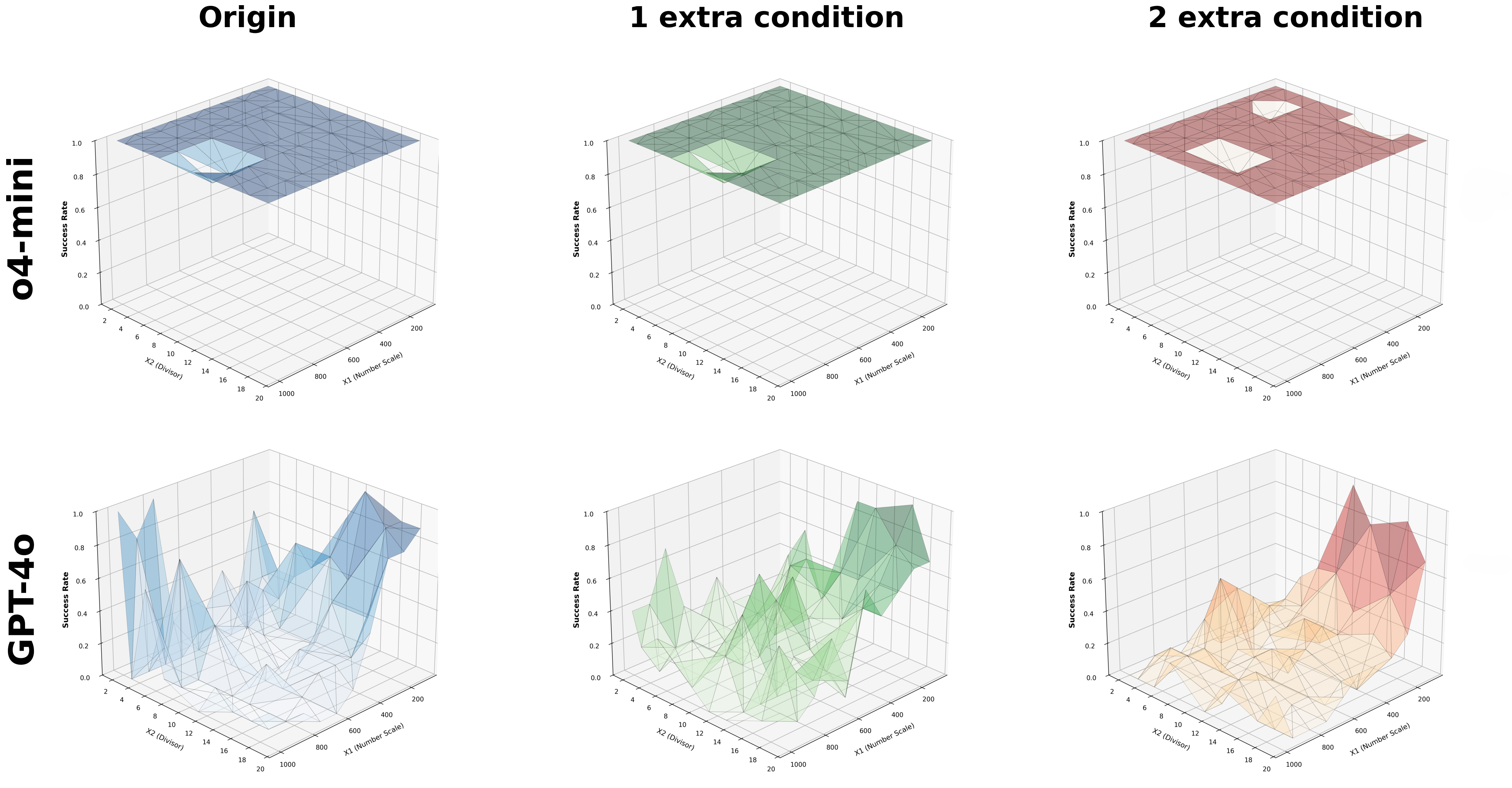}
    \caption{Capability of \textsc{GPT-4o} and \textsc{o4-mini} measured on N-primable problems as a constraint refinement example. The model's performance remains relatively unchanged when refining constraints are introduced.}
    \label{fig:motivating_0}
\end{figure}

\paragraph{Capability Frontier.}
As illustrated in Figure~\ref{fig:motivating_0}, the performance landscape reveals a striking divergence in model behavior under structural pressure. 
\textsc{o4-mini} maintains a stable, high-success \emph{plateau} across all coordinates, indicating that its reasoning capacity is largely invariant to both the expansion of the search space and the injection of compositional constraints. 
In contrast, \textsc{GPT-4o} demonstrates a heightened sensitivity to the parameters expanding. 
While it maintains acceptable performance in regions with a small, highly constrained solution space, its success rate becomes volatile as the search space expands. 
Notably, the transition from one to two additional conditions does not induce a monotonic collapse in either model.
This phenomenon indicates that adding conditions such as \textit{non\_decreasing} primarily refines the constraints and filters the existing solution set rather than fundamentally altering the underlying solution topology. 
Models are relatively robust to \textit{constraint refinement}, in which additional structural constraints are added, just shrinking the existing solution space.
For this class of problems, the key challenge is whether LLMs can be \textbf{more reliable} in producing the correct answer for a certain problem in 10 trials.
The \textsc{o4-mini} shows a higher reliability than \textsc{GPT-4o}.

\subsection{Postage Stamp Problem}
\label{sec:motivating}
Consider another question:
\emph{``Using X2 stamps of denominations {1, 5}, what is the smallest number of {5}-cent stamps required to form every integer amount from 1 to X1?''}

\paragraph{Autoformalization.}
The autoformalizer and the parameterizer conduct the same transformation. 
For two denominations $\{d_1,d_2\}$, it introduces decision variables
$\texttt{cnt}[i]$ denoting how many stamps of denomination $d_i$ are available,
and $\texttt{use}[t][i]$ denoting how many stamps of denomination $d_i$ are used to form the target amount $t$.
It enforces \emph{coverage} by requiring that every target amount admits a feasible
sub-multiset of the available stamps.
The pseudo code is:
\begin{verbatim}
 for each t in [1, X1]:
   exists use[t][i] s.t. 0 <= use[t][i] <= cnt[i]
   sum_i use[t][i] * d[i] == t
sum_i cnt[i] = X2
...
\end{verbatim}

\paragraph{Structural Calibration.}
% The well-known postage stamp problem~\cite{10.1145/507457.507473} asks which consecutive integer amounts can be formed using a bounded number of stamps from a fixed denomination set. 
% Given denominations \(D=\{d_1,\ldots,d_m\}\) and a stamp budget \(h\), the classical objective is to find the largest \(N\) such that every amount \(t\in\{1,\ldots,N\}\) admits a representation
% \[
% t=\sum_{i=1}^{m} a_i d_i,
% \qquad 
% a_i\in\mathbb{Z}_{\ge 0},
% \qquad
% \sum_{i=1}^{m} a_i\le h.
% \]
% The above example is a coverage variant in which the total number of available stamps is fixed, and the objective is to minimize the number of stamps of a specified denomination while ensuring that every amount in a target interval can still be represented.
Similarly, \tool can generate the variants of the example. 
For example,
\emph{``Using denominations $\{1,d_1\}$ with $1<d_1$, and requiring at least one stamp of each denomination, what are the three smallest values of $d_1$ for which the minimum total number of stamps required to form every integer amount from $1$ to $X1$ is exactly $X2$?''}
Importantly, this is not a surface paraphrase, and the solution manifold is totally changed.
$X_1$ and $X_2$ directly control the coverage requirement and resource budget, thereby inducing monotone changes in feasibility and optimality.
Similarly, \tool can further generate a new problem with three types of face value: 
\emph{``Using denominations $\{1,d_1,d_2\}$ with $1<d_1<d_2$, and requiring at least one stamp of each denomination, what are the three smallest values of $d_2$ for which the minimum total number of stamps required to form every integer amount from $1$ to $X1$ is exactly $X2$?''}

\begin{figure}
    \centering
    \includegraphics[width=1\linewidth]{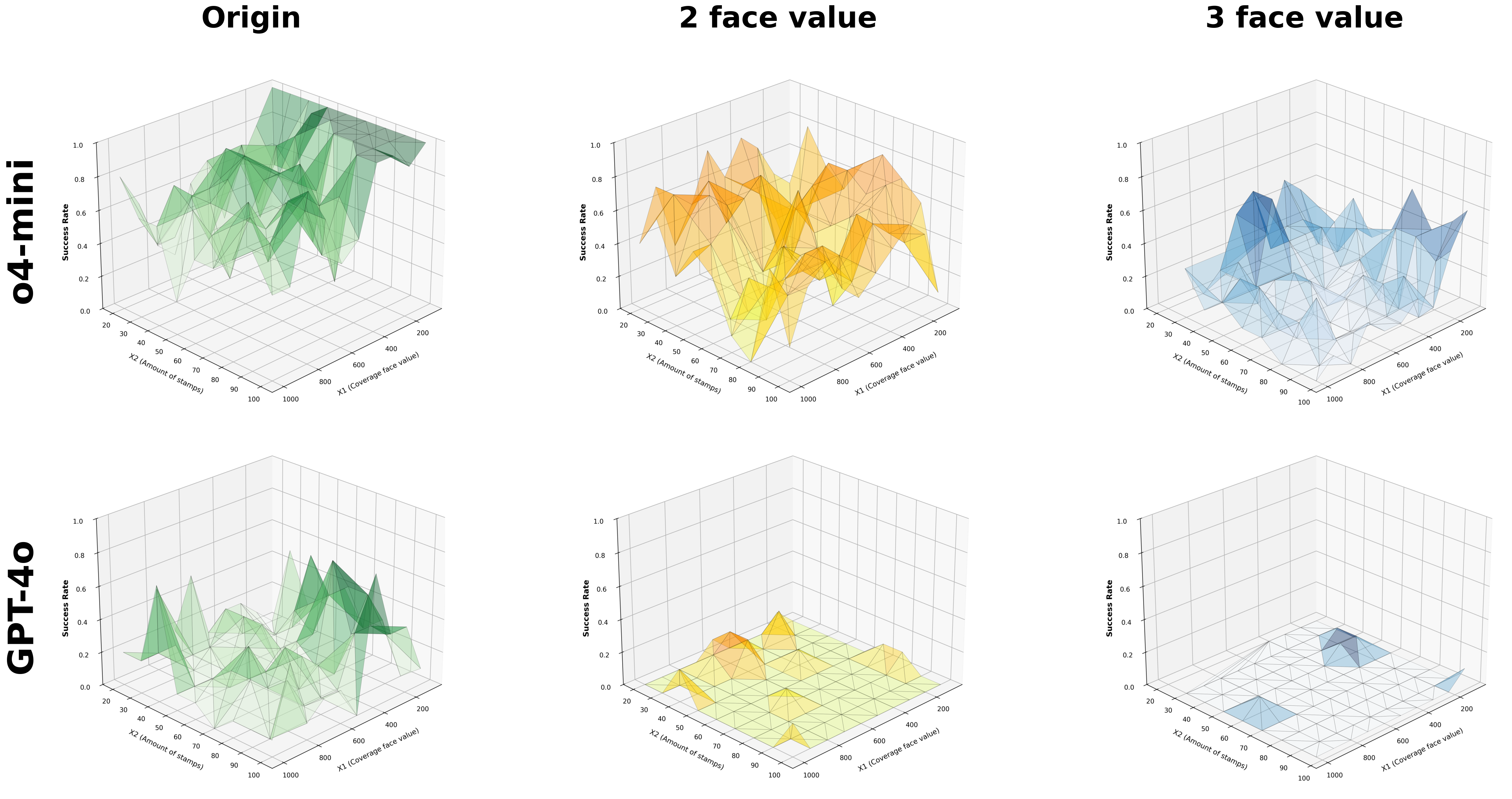}
    \caption{Capability of \textsc{GPT-4o} and \textsc{o4-mini} measured on the stamp-coverage problem as a solution space restructuring example. The models' performance is decreasing sharply when the calibration restructures the solution manifold.}
    \label{fig:motivating}
\end{figure}

\paragraph{Probing and Capability Frontier.}
% \tool randomizes $(X_1, X_2)$ within bounded domains to generate a batch of solver-verified probe instances, and evaluates the LLMs on these instances following the same protocol as in the motivating example.
\autoref{fig:motivating} visualizes the empirically observed capabilities of \textsc{GPT-4o} and \textsc{o4-mini} under structural parameterizations.
As the structural complexity increases, the valid region contracts monotonically for both models. 
However, the contraction is substantially sharper for \textsc{GPT-4o}. 
Its performance remains relatively stable when the structural variation only induces incremental constraint refinement, but deteriorates rapidly once the task requires a non-trivial reorganization of the solution space. 
In contrast, \textsc{o4-mini} maintains a much broader stable region, suggesting greater robustness to structural changes.
This pattern is not merely a smooth degradation in accuracy. 
Instead, it reveals a pronounced \emph{transition regime}: models can remain stable under refinement-like perturbations, yet fail abruptly when the same problem family crosses into a restructuring-sensitive region. 
We interpret this transition as evidence of a capability frontier, where structural parameterization exposes the boundary between reasoning that is stable under constraint refinement and reasoning that requires substantive restructuring of the underlying solution representation.
The sharp separation between \textsc{GPT-4o} and \textsc{o4-mini} further suggests that such capability frontiers capture qualitative differences in structural reasoning capacity, which would be obscured by aggregate accuracy metrics alone.

\subsection{Chain-of-Thought Structural Reasoning}
\label{sec:cot_training}
The question is
\emph{``Two ice pucks collide head-on on a frictionless surface. Puck A has mass $0.691\,\mathrm{kg}$ and moves at $3\,\mathrm{m/s}$ toward a stationary puck B of mass $1\,\mathrm{kg}$. After the collision, puck A reverses direction and moves at $3\,\mathrm{m/s}$ opposite its original direction. Determine the impulse on puck A and on puck B.''}
Although this instance is arithmetically simple, it exposes a common failure mode of chain-of-thought reasoning:
the model can compute a locally plausible quantity while failing to recompose the result under a global structural invariant.
The impulse on each puck must equal its own change in momentum,
$
J_i = \Delta p_i,
$
and the two impulses must satisfy the coupled system-level constraint
$
J_A + J_B = 0,
$
which follows from Newton's third law, or equivalently from conservation of momentum for the isolated two-puck system.
Thus, a correct solution must satisfy both local numerical constraints and a global coupling constraint.
In our framework, the probe is decomposed into atomic, machine-checkable reasoning states:
extracting masses and velocities, computing each momentum change, checking signs and units, and enforcing the global impulse-coupling constraint before recomposing the final answer.
Each intermediate state can be verified by executable formal code, producing supervision that identifies not only whether the final answer is wrong, but also which structural constraint was violated.
% More details are shown in \autoref{app_moti}.
% This example illustrates how formalized probes can support training, rather than only evaluation.
More importantly, this supervision scales naturally to structurally varied probes.
The same invariant-coupling pattern can be calibrated along axes such as the number of interacting objects, the number of conservation laws, the depth of dependency chains, and the degree of cross-constraint coupling.
As these structural parameters increase, models may continue to succeed on local arithmetic while failing at specific recomposition points.
The resulting probe outcomes, therefore, reveal where a model's reasoning breaks down in the structured problem space.
% Such information can be used to construct a targeted training curriculum.
Instead of simply encouraging longer chains of thought, \tool provides verified intermediate supervision for structural reasoning.
% Given a problem, \tool identifies the structural invariant required to combine local results into a globally consistent solution, generates sub-questions that isolate this invariant, and verifies each step against executable constraints.
% Training on these traces teaches the model when an intermediate numerical answer is insufficient, which global invariant must be invoked, and how the final answer should be repaired when local computations conflict with system-level constraints.
% In this way, formalized and calibrated probes serve two complementary roles.
% First, they expose reasoning failures that are hidden beneath fluent chain-of-thought explanations.
% Second, they provide structured, verifiable supervision for improving the model's ability to handle increasing structural difficulty.
% The goal is therefore not to lengthen the reasoning trace, but to strengthen the model's ability to extract, maintain, and recompose constraints across structurally complex reasoning tasks.

 % 1.5
\section{Approach}
\begin{figure}
    \centering
    \includegraphics[width=0.9\linewidth]{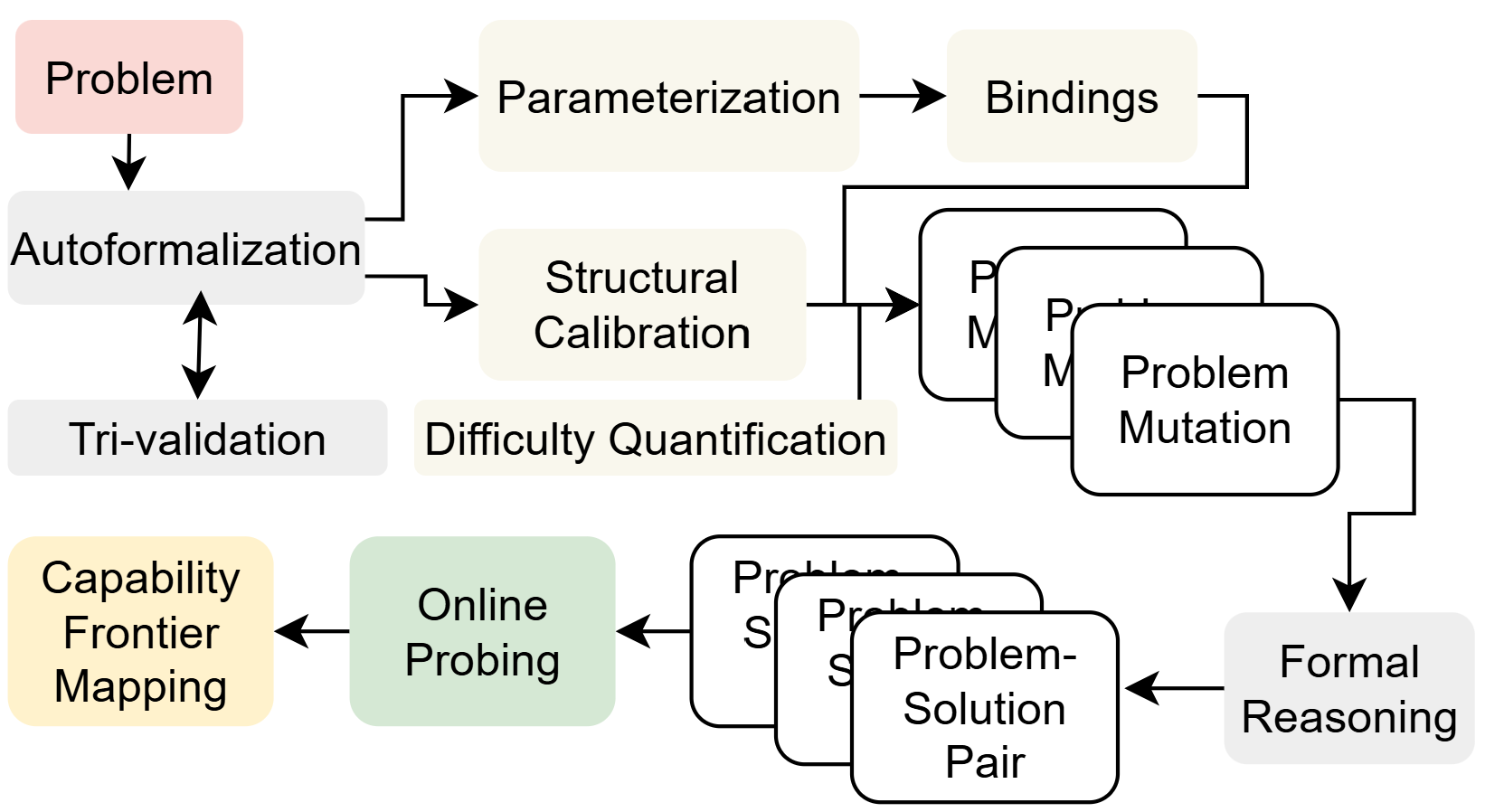}
    \caption{\tool Framework Overview.}
    \label{fig:framework}
\end{figure}

% from example-level to dataset analysis
% from specific case to general probes
\label{sec:approach}
We propose \textbf{\tool} (e\textbf{X}plainable \textbf{R}easoning \textbf{A}nalysis, s\textbf{Y}stem), 
a unified evaluation framework consisting of five tightly coupled components shown in \autoref{fig:framework}:
(1) autoformalization,
(2) difficulty quantification,
(3) controlled calibration,
(4) formal verification, and
(5) online probing and capability frontier mapping.

\subsection{Autoformalization}
\label{sec:autoformalization}

The first step is to transform natural-language reasoning tasks into explicit, executable representations.
Given a natural-language problem description $\mathcal{P}_{\text{NL}}$,
\tool employs an LLM-based autoformalizer to generate a corresponding formal artifact  $\mathcal{P}_{\text{Code}}$ that encodes the solution logic.
The artifact may target different formal reasoning backends (e.g., SMT, theorem provers, or symbolic algebra).
but should satisfy three requirements:
(i) \emph{semantic completeness}, capturing all constraints required to solve the task;
(ii) \emph{executability}, enabling solver-based reasoning; and
(iii) \emph{traceability}, allowing alignment between natural-language entities and formal variables.
We make this alignment explicit by defining a binding map
$
\mathcal{B} : \mathcal{V}_{\text{NL}} \;\leftrightarrow\; \mathcal{V}_{\text{Code}},
$
where $\mathcal{V}_{\text{NL}}$ denotes identifiable variables in the natural-language prompt and
$\mathcal{V}_{\text{Code}}$ the corresponding formal variables.
Beyond enabling solver-based execution, autoformalization serves as a canonicalization step that collapses diverse surface realizations into a shared structural representation.
Consequently, probes are compared, calibrated, and verified at the level of formal structure rather than linguistic form.
%%%%
% \paragraph{Tri-validation.} 
Validation for the autoformalizing is implemented as a composite operator
$
\mathcal{T} =
\mathcal{T}_{\text{static}} \circ
\mathcal{T}_{\text{dynamic}} \circ
\mathcal{T}_{\text{semantic}},
$
where $\mathcal{T}_{\text{static}}$ checks syntactic well-formedness and typing,
$\mathcal{T}_{\text{dynamic}}$  executes the artifact on randomized instances and compares outputs across independent formal backends when
available,
and $\mathcal{T}_{\text{semantic}}$ employs an auxiliary LLM to audit alignment between the formal encoding and the intended natural-language semantics with the binding map $\mathcal{B}$.
Only probes that satisfy $\mathcal{T}=\mathsf{true}$ are admitted into the evaluation set.

\subsection{Difficulty Quantification}
\label{sec:difficulty}

To meaningfully order probes, we define difficulty in terms of the \emph{structural information} encoded in the formal specification, rather than empirical model performance.
Intuitively, difficulty reflects the amount of structured information that must be simultaneously extracted, composed, and maintained to reach a correct solution.
Each probe is associated with a structural descriptor
$
\boldsymbol{\theta} = (c, d, \kappa, \ell),
$
where
$c$ denotes \emph{conjunctive width}, measuring how many constraints must be satisfied simultaneously;
$d$ denotes \emph{compositional depth}, induced by nesting, branching, or conditional structure;
$\kappa$ captures \emph{cross-constraint coupling} through shared variables or derived quantities;
and $\ell$ measures the minimal \emph{dependency-chain length} required to derive the target output.
% Together, these quantities characterize the amount of extractable structure independent of surface realization.
It is computed directly from the formal artifact $\mathcal{P}_{\text{Code}}$.
%
% \paragraph{From example to benchmark.}
We also compute solver-grounded complexity measures,
including expression size $E_{\text{expr}}$, reasoning depth $E_{\text{reason}}$,
and solver resource usage $(E_{\text{space}}, E_{\text{time}})$.
While these quantities characterize the intrinsic difficulty of the formal problem, they are not directly observable by language models.
Our structural descriptor can be viewed as an abstraction of these quantities projected onto the space of structures accessible to LLMs,
enabling difficulty to be treated as an explicit and controllable variable.

\subsection{Controlled Calibration}
To isolate structural effects and mitigate contamination, we apply controlled calibration via a family of transformations.
Concretely, constants in $\mathcal{P}_{\text{Code}}$ are replaced with symbolic variables
$\{x_1,\dots,x_n\}$ drawn from bounded domains $\mathcal{D}_{x_i}$ that respect type and semantic invariants:
% \[
% x_i \sim \mathcal{D}_{x_i} \subseteq \mathcal{T}(x_i),
% \quad \text{s.t. } \phi(x_1,\dots,x_n).
% \]
Such substitutions alter surface realizations without modifying the underlying constraint graph or its dependency relations.
We randomize the parameters, such as numeric bounds or thresholds.
These transformations affect task size but do not introduce new constraints or dependency paths,
and are therefore treated as an auxiliary calibration axis.
To further make structural difficulty explicitly controllable, we introduce a compositional intermediate representation that exposes the internal structure of each probe as an object amenable to formal transformation.
Each probe $\mathcal{P}$ is represented as a tuple
$
\mathcal{I} =
(\mathcal{P}_{\text{Code}}, \mathcal{P}_{\text{NL}}, \mathcal{B}, \mathcal{C}, \mathcal{S}),
$
where: $\mathcal{C}$ is a finite set of constraints in the probe, and $\mathcal{S}$ is a structural skeleton capturing compositional form (e.g., sequential composition, nesting depth, or multi-step derivation chains).
%%%%%%%%%%%
% \paragraph{Structural Operators.}
We define the structure-transforming operators into two classes.
\emph{(1) Constraint refinement operators.}
These operators increase conjunctive width or tighten solution regions
without altering the global dependency topology, including
constraint conjunction: replacing $\mathcal{C}$ with $\mathcal{C} \cup \{c_i\}$,
constraint tightening: replacing $\mathcal{C}$ with $ \mathcal{C} \cap \{c_i\}$,
domain restriction: shrinking $\mathcal{S}$ while preserving type invariants.
\emph{(2) Structural restructuring operators.}
These operators alter compositional topology or dependency geometry, including
nesting introduction: embedding the code within a conditional or iterative structure,
cross-variable coupling: introducing shared derived quantities,
dependency chaining: inserting intermediate latent variables,
representation shift: replacing a direct constraint with a derived multi-step formulation.
% These transformations modify $\mathcal{C}$ or $\mathcal{S}$,
% affecting depth $d$, coupling $\kappa$, or dependency length $\ell$, and induce non-linear changes in the demand for reasoning.

\subsection{Formal Reasoning}
\label{sec:verification}

Formal reasoning enforces correctness and well-posedness prior to evaluation,
serving as the foundation for reliable capability measurement.
For each instantiated probe, we compute a canonical answer using formal reasoning engines.
% and require that the specification is well-defined.
In particular, we enforce the existence of one solution and uniqueness of the solution:
\begin{align*}
\text{(Existence)} \quad &
\exists solution\ y : \mathcal{P}_{\text{Code}} \rightarrow y, \\
\text{(Uniqueness)} \quad &
\forall y_1,y_2,\;
(\mathcal{P}_{\text{Code}} \rightarrow y_1 \land
 \mathcal{P}_{\text{Code}} \rightarrow y_2)
\Rightarrow y_1 = y_2.
\end{align*}
By enforcing existence and uniqueness \emph{before} evaluation, we ensure that each probe corresponds to a
valid and unambiguous measurement point in the structured probe space, preventing ambiguity from
blurring capability boundaries.

\subsection{Probing and Capability Frontier}
\label{sec:evaluation}

Online evaluation presents calibrated probes to a target LLM and compares model predictions $\hat{y}$
against canonical answers $y^\star$ across multiple randomized instantiations.
For each probe family, we traverse the structured probe space by systematically increasing
$\boldsymbol{\theta}$ along one or more structural axes.
Each structural configuration is replicated using controlled calibration to ensure that
observed performance differences are attributable to structural variation rather than
surface realization.
We define a capability frontier as a structural property of a model over a calibrated probe space, rather than as the outcome of an isolated instance or an aggregate dataset score.
Let
$
\mathcal{X}=\{x(\boldsymbol{\theta},r)\mid \boldsymbol{\theta}\in\Theta,\ r\in\mathcal{R}(\boldsymbol{\theta})\}
$
denote the probe space, where \(\boldsymbol{\theta}\) specifies structural parameters and \(r\) indexes randomized surface realizations.
For a model \(M\), we define its local reliability at structure \(\boldsymbol{\theta}\) as
\[
p_M(\boldsymbol{\theta})
=
\Pr_{r\sim\mathcal{R}(\boldsymbol{\theta})}
\left[
M(x(\boldsymbol{\theta},r))=y^\star(\boldsymbol{\theta},r)
\right].
\]
Given a reliability threshold \(\tau\), the capability frontier is the transition region in \(\Theta\) where \(p_M(\boldsymbol{\theta})\) changes from stable high reliability to unreliable performance as structural difficulty increases.
Formally, for a degradation margin \(\Delta>0\), we define
\[
\mathcal{F}_{M,\tau,\Delta}
=
\left\{
\boldsymbol{\theta}\in\Theta
\;\middle|\;
p_M(\boldsymbol{\theta})\ge \tau
\ \land\
\exists \boldsymbol{\theta}'
\text{ such that }
p_M(\boldsymbol{\theta}')\le \tau-\Delta
\right\}.
\]
Intuitively, the frontier marks the region of stable reliability before structural difficulty induces a substantial performance collapse, distinguishing genuine capability limits from minor fluctuations.
% Thus, individual instances serve only as samples for estimating local reliability, while datasets are useful insofar as they provide calibrated coverage of the structured probe space. % 1.5
\section{Experiments}
\label{sec:experiments}
% 实验部分的解释还可以再精简
Our experiments are guided by the following research questions.

\textbf{RQ1}: \textit{How does model performance vary across a structured difficulty space?}
We study how LLM success rates change as formal problem structure varies along dimensions 
such as expression complexity and proof depth.

\textbf{RQ2}: \textit{Do different models exhibit distinct capability and phase transition behaviors?}
We compare multiple LLMs and reasoning-specialized models to analyze whether they fail in similar regions or exhibit qualitatively different capability geometries.

\textbf{RQ3}: \textit{Can solver-verified structural supervision systematically improve LLM's reasoning capability?}
We perform controlled fine-tuning with solver-verified Chain-of-Thought and evaluate whether model performance moves consistently and measurably within the structured difficulty space.

\textbf{RQ4}: \textit{How reliable is \tool in generating valid and semantically faithful probes?}
We assess whether generated probes pass tri-verification, including compilation into executable code, dynamic validation, and LLM-based judgment.

\subsection{Experimental Settings}
\textbf{Datasets.}
We sample seed problems in the following datasets:
\textbf{GSM8K}~\citep{cobbe2021gsm8k} is a dataset of grade-school math word problems requiring multi-step reasoning. 
\textbf{MATH}~\citep{hendrycks2021math} is a more challenging dataset covering advanced topics in high school mathematics.
\textbf{PHYSICS}~\citep{pocket-physics2023} is a public dataset of high school-level physics problems. 
\textbf{CHEMISTRY}~\citep{wei2021chemistryqa} is a public dataset of high school-level chemistry problems. 
We automatically and successfully formalize 5876 problems in the GSM8K dataset. 
For the other datasets, we generate 1,000 samples from each to demonstrate the generalizability of our framework across domains and problem types.
% Our framework is scalable, and larger formalized datasets can be constructed in future work with increased resources and funding.

\noindent \textbf{Baselines.}
We tested \tool with several state-of-the-art LLMs including GPT-5~\citep{GPT5}, o4-mini~\citep{o4mini}, GPT-4o~\citep{achiam2023gpt}, Qwen-Plus-2025-04-28~\citep{bai2023qwen}, Qwen2-MATH~\citep{qwen25math}, QwQ~\citep{qwq32b}, Claude-3.5 Sonnet~\citep{anthropic2024claude}, DeepSeek-V3~\citep{liu2024deepseek}.
We employ Z3~\citep{de2008z3}, CVC5~\citep{barbosa2022cvc5}, and Mathematica 14.0~\citep{mathematica14} for formal validation. 

\noindent \textbf{Implementation Details.}
% 校对一下这里 %%%%%%%%%%%%%%%%%%%%%%%%%%%%%%%%%%%%%%%%%%%%%%%%%%%%%%%%%%%%%%%
We selected three models as our backbones for the training experiment: DeepSeek-R1-1.5B-Distill, GLM-4.1V-9B-Thinking, and Qwen3-14B-Thinking.
We conducted all fine-tuning experiments using the LLaMA-Factory framework. 
To ensure a fair comparison, we maintained a unified hyperparameter configuration across all backbones.
% We show the details on the website \citep{ace-nips}. 
% More experiment results are in the appendix \autoref{tab:math_domains} and in \autoref{tab:science_domains}.
% The detailed token and time usage are in \autoref{tab:token_usage}.
The models were fine-tuned using LoRA for parameter efficiency. Training was performed for 30 epochs using the AdamW optimizer with an initial learning rate of 5e-5 and a cosine learning rate scheduler. We utilized BF16 precision to optimize computational throughput and stability.
Given the memory-intensive nature of CoT training, we employed a gradient accumulation strategy. We set the per-device batch size to 4 and accumulated gradients over 8 steps. A maximum gradient norm of 1.0 was applied for gradient clipping.
% \paragraph{Hardware and Environment}
All experiments were conducted on NVIDIA RTX 6000 Ada Generation GPUs (48GB VRAM). The software environment was configured with CUDA 12.4 and PyTorch 2.9.1.

\subsection{RQ1: Performance Probing}
Tables~\ref{tab:all_domains_avg} show that formally calibrated probes consistently expose robustness gaps across mathematical and scientific reasoning domains.
Although many models achieve high original accuracy, their performance drops substantially after randomization.
This indicates that original benchmark performance can overestimate the stability of model reasoning under structural variation.
The drop is particularly informative because it separates models that appear similar under the original evaluation.
For example, most models obtain near-saturated original accuracy on GSM8K, but their randomized averages range from 66.34\% to 80.51\%.
Similarly, in the science domains, Claude-3.5 and Qwen-Plus achieve strong original scores but suffer large degradation gaps, while GPT-5 and o4-mini maintain much smaller gaps.
This suggests that randomized variants reveal differences in structural robustness that are hidden by original benchmark scores.
%%%%%%%
Statistical analysis confirms that the observed degradation is systematic.
The degradation is statistically significant across all four domains:
GSM8K shows an average drop of 25.12 points ($p<0.001$),
MATH shows an average drop of 30.50 points ($p<0.001$),
PHYSICS shows an average drop of 26.01 points ($p=0.0014$),
and CHEMISTRY shows an average drop of 23.14 points ($p=0.0028$).
% Paired tests across models show significant drops from original accuracy to randomized accuracy on GSM8K, MATH, PHYSICS, and CHEMISTRY, with average decreases of 25.12, 30.50, 26.01, and 23.14 points, respectively.
All Wilcoxon signed-rank tests are also significant, showing that the effect is consistent across models rather than caused by a few outliers.

\begin{table*}[t]
\centering
\footnotesize
\caption{Performance of models across four domains. Orig denotes the model performance on the original datasets. Avg denotes the mean over three randomized variants on \tool benchmarks, and Std denotes the standard deviation across them.}
\label{tab:all_domains_avg}
\resizebox{\textwidth}{!}{%
\begin{tabular}{l|ccc|ccc|ccc|ccc}
\toprule
\multirow{2}{*}{\textbf{Model}} &
\multicolumn{3}{c|}{\textbf{GSM8K}} &
\multicolumn{3}{c|}{\textbf{MATH}} &
\multicolumn{3}{c|}{\textbf{PHYSICS}} &
\multicolumn{3}{c}{\textbf{CHEMISTRY}} \\
\cmidrule(lr){2-4} \cmidrule(lr){5-7} \cmidrule(lr){8-10} \cmidrule(lr){11-13}
& Orig & Avg$\pm$Std & Gap
& Orig & Avg$\pm$Std & Gap
& Orig & Avg$\pm$Std & Gap
& Orig & Avg$\pm$Std & Gap \\
\midrule
Claude-3.5
& 98.05 & 71.30$\pm$1.07 & 26.75
& 85.35 & 60.89$\pm$1.99 & 24.46
& \textbf{88.91} & 45.46$\pm$2.18 & 43.45
& \textbf{89.72} & 49.22$\pm$0.72 & 40.50 \\

DeepSeek-V3
& 98.10 & 71.27$\pm$1.27 & 26.83
& 98.17 & 64.10$\pm$2.02 & 34.07
& 79.71 & 56.70$\pm$3.51 & 23.01
& 87.49 & 59.67$\pm$1.28 & 27.82 \\

GPT-4o
& 97.44 & 66.34$\pm$0.64 & 31.10
& 85.16 & 57.02$\pm$1.75 & 28.14
& 76.10 & 43.79$\pm$1.47 & 32.31
& 81.38 & 43.86$\pm$0.95 & 37.52 \\

o4-mini
& \textbf{98.66} & \textbf{80.51}$\pm$3.01 & 18.15
& 98.71 & 69.11$\pm$2.02 & 29.60
& 79.56 & 69.41$\pm$1.26 & 10.15
& 80.19 & \textbf{75.80}$\pm$0.85 & 4.39 \\

GPT-5
& 98.35 & 80.00$\pm$0.83 & 18.35
& \textbf{99.72} & \textbf{72.58}$\pm$0.23 & 27.14
& 76.38 & \textbf{69.85}$\pm$1.06 & 6.53
& 79.60 & 72.29$\pm$1.06 & 7.31 \\

Qwen-Plus
& 98.40 & 68.81$\pm$1.71 & 29.59
& 92.31 & 60.62$\pm$1.14 & 31.69
& 88.27 & 42.06$\pm$2.28 & 46.21
& 79.97 & 44.28$\pm$0.54 & 35.69 \\

Qwen2-MATH
& 98.13 & 68.53$\pm$2.28 & 29.60
& 91.94 & 63.21$\pm$0.87 & 28.73
& 69.09 & 41.28$\pm$1.23 & 27.81
& 66.71 & 43.88$\pm$1.26 & 22.83 \\

QwQ
& 98.56 & 77.95$\pm$1.65 & 20.61
& 97.00 & 56.84$\pm$7.81 & 40.16
& 79.23 & 60.64$\pm$0.99 & 18.59
& 83.69 & 74.65$\pm$0.52 & 9.04 \\
\bottomrule
\end{tabular}
}
\end{table*}

\subsection{RQ2: Capability Geometries of LLMs}
\label{sec:rq2}
\begin{figure}[htbp]
    \centering
    \includegraphics[width=0.9\linewidth]{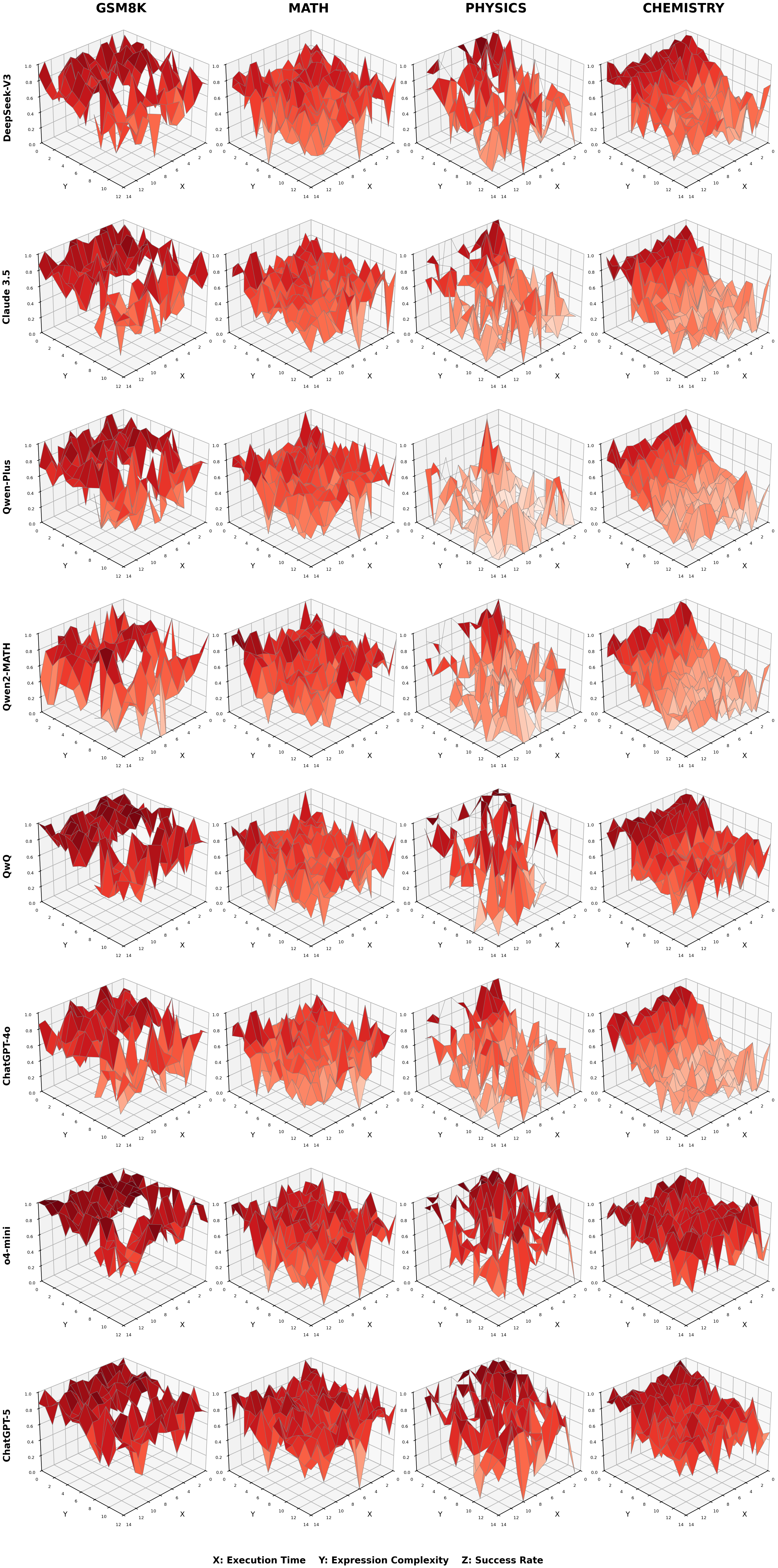}
    \caption{Capability surfaces of model success rates over execution time and expression complexity across four datasets.}
    \label{fig:final}
\end{figure}

We continuously scale problem difficulty along formal structural dimensions
and measure model success rates over the resulting difficulty grid.
Figure~\ref{fig:final} visualizes model performance as a capability surface over two structural axes: execution time and expression complexity.
For each dataset--model pair, we construct six two-dimensional heatmaps
corresponding to every pairwise combination of the four structural
dimensions including execution Time, expression complexity, reasoning Depth, and
state-Space size, yielding an $8 \times 6$ matrix of capability portraits shown in \autoref{appendix}.

% (Figures~\ref{fig:combined_gsm8k}--\ref{fig:combined_chemistry}).

% \subsubsection{Cross-Model Observations}
% 物理：o4-mini and ChatGPT-5 > qwq and deepseek-V3 > claude 3.5 and qwen2-math and chatgpt-4o > qwen-Plus
% 化学：o4-mini and ChatGPT-5 > deepseek-V3 > qwq > claude 3.5 and qwen2-math and chatgpt-4o and qwen-Plus
% GSM8K: o4-mini and chatgpt-5 更深一些，其他差不多
% MATH: o4-mini and ChatGPT-5 > qwq and qwen2math > 其他
\subsubsection{Cross-Model Observations}

Across the four benchmarks, \textsc{GPT-5} and \textsc{o4-mini} consistently exhibit the strongest structural robustness.
They maintain dense and uniform high-performance regions across all six pairwise structural projections, indicating that their success is not limited to a particular difficulty axis.
In both arithmetic-oriented benchmarks and domain-specific scientific benchmarks, these two models are better able to preserve accuracy when reasoning depth is coupled with expression complexity, execution time, or state-space size.
The second-tier models show more domain-dependent behavior.
In Physics, \textsc{QwQ} and \textsc{DeepSeek-V3} form a clear middle group: they outperform \textsc{Claude-3.5}, \textsc{Qwen2-MATH}, and \textsc{GPT-4o}, but still exhibit more fragmented frontier regions than \textsc{o4-mini} and \textsc{GPT-5}.
In Chemistry, \textsc{DeepSeek-V3} remains the strongest model below the top tier, followed by \textsc{QwQ}.
This suggests that \textsc{DeepSeek-V3} is relatively more robust in domain-specific settings that require maintaining symbolic quantities, state dependencies, and scientific constraints.
For the math-oriented benchmarks, the model ordering is slightly different.
On GSM8K, most models achieve broadly similar performance, while \textsc{o4-mini} and \textsc{GPT-5} still show deeper and more uniform high-accuracy regions.
This indicates that GSM8K is close to saturated for current models, making cross-model differences less visible except under coupled structural perturbations.
On MATH, the separation becomes clearer: \textsc{o4-mini} and \textsc{GPT-5} remain the strongest, followed by \textsc{QwQ} and \textsc{Qwen2-MATH}, while the remaining models show more localized instability.
% %
% Overall, the cross-model comparison shows that stronger models are not merely higher in aggregate accuracy, but have more stable and contiguous capability frontiers.
% Weaker models may perform well on isolated structural dimensions, but their valid regions become sparse once multiple dimensions interact.
% This reinforces the central finding that model capability should be characterized by the shape of its structural frontier, rather than by a single averaged score.

\subsubsection{Cross-Domain Observations}

The four benchmarks form a clear hierarchy of difficulty based on LLM capabilities.
% on GSM8K (Figure~\ref{fig:combined_gsm8k}), nearly every cell across all
% models is saturated deep red, indicating near-ceiling accuracy regardless of
% the structural combination.  On MATH (Figure~\ref{fig:combined_math}), the
% heatmaps remain predominantly red but high-dimensional corners begin to
% bleach noticeably.  
% Chemistry (Figure~\ref{fig:combined_chemistry})
% introduces further degradation, and Physics
% (Figure~\ref{fig:combined_physics}) exhibits the most widespread accuracy
% loss, with multiple dimension pairs showing large pale or missing regions.
The overall ordering is
$
  \text{GSM8K} < \text{MATH} < \text{Chemistry} < \text{Physics}
$
in terms of LLM capabilities on structural reasoning.
Further analysis shows that the drop from MATH to Physics is far steeper than the drop from
GSM8K to MATH: the former involves a qualitative shift.
Physical problems demand situational modeling and causal grounding on top of formal
manipulation rather than a mere quantitative increase in difficulty.
Chemistry occupies an intermediate position: it requires domain-specific
knowledge (reaction pathways, stoichiometry, molecular structure) but relies
less on extended causal chains than physics does.

\subsubsection{Cross-structure Observations}

Across all domains, the cross-structure heatmaps reveal that model capability frontiers are not determined by a single structural axis, but by interactions among multiple dimensions of problem structure.
Single-axis growth, such as increasing execution time, expression complexity, or state-space size alone, often leads to relatively mild or non-monotonic degradation.
In contrast, sharper performance transitions emerge when these dimensions are coupled with reasoning depth.
This pattern is consistently observed in the \emph{Depth--Complexity}, \emph{Depth--Space}, and \emph{Depth--Time} projections, where models must not only process larger structures but also maintain and compose intermediate dependencies across multiple reasoning steps.
The effect of such structural coupling varies substantially across domains.
GSM8K exhibits the most saturated landscape: most models remain stable across a wide range of structural configurations, and failures appear mainly as localized holes rather than large collapse regions.
MATH is slightly more fragmented, especially under depth-related projections, but still remains broadly stable due to its alignment with symbolic reasoning patterns commonly seen during pretraining.
By contrast, Physics and Chemistry expose much clearer capability frontiers.
In these domains, structural growth often requires tracking system-level constraints, units, conservation laws, reactions, quantities, or state-dependent dependencies.
As a result, failures become more frequent and more discontinuous once reasoning depth is combined with expression complexity or state-space variation.

The comparison across models further shows that stronger reasoning models preserve larger and more contiguous high-performance regions, while weaker or less specialized models exhibit sparse and fragmented valid regions under coupled structural pressure.
Notably, these differences are not fully captured by aggregate accuracy.
Two models with similar average performance may exhibit different frontier shapes: one may remain robust under shallow but large state spaces, while another may fail primarily when depth forces multi-step recomposition.
Thus, the heatmaps characterize not only how often a model succeeds, but also which structural interactions trigger its failures.
A recurring pattern across domains is the absence of smooth monotonic degradation.
Instead of gradually declining as a scalar difficulty score increases, model performance often remains stable within a region and then collapses abruptly for specific structural combinations.
This supports the view that LLM reasoning failures are interaction-driven: brittleness appears when models must jointly extract, maintain, and recombine multiple dependent structures, rather than when any single dimension becomes large in isolation.

% Overall, these results demonstrate that capability frontiers are inherently multi-dimensional and domain-dependent.
% Arithmetic-oriented benchmarks such as GSM8K and MATH mostly test whether models can sustain regular symbolic or numerical chains, whereas Physics and Chemistry additionally require domain-specific structural consistency.
% By exposing these differences, \tool{} provides a fine-grained diagnostic framework that goes beyond aggregate benchmark scores and identifies the precise structural regimes in which model reasoning remains stable or breaks down.

% \subsubsection{Checkerboard Instability in Reasoning Models}

A striking visual signature distinguishes certain models: alternating
dark--light \emph{checkerboard} textures in which adjacent difficulty bins
exhibit sharply different accuracy.  This pattern is most prominent in QwQ
and, to a lesser extent, o4-mini.  On GSM8K the checkerboard is virtually
absent; on MATH it appears mildly; on Chemistry it intensifies (especially
in the Time\;vs.\;Complexity and Time\;vs.\;Space columns); and on Physics
it becomes severe.
We interpret this instability as evidence that chain-of-thought reasoning
strategies are \emph{brittle with respect to small perturbations in problem
structure}: certain parameter combinations happen to align with the model's
reasoning templates and succeed, while nearby combinations fall into blind
spots and fail.  As task difficulty increases, the probability of any single
reasoning step going astray grows, amplifying the oscillatory behaviour.
In contrast, the strongest models, GPT-5, display
markedly smoother, more uniform color distributions across all dimensions
pairs, indicating lower sensitivity to structural variation.  This
\emph{uniformity of the capability surface}, rather than peak accuracy on
any single slice, may be the most reliable indicator of robust reasoning
capability.

\subsection{RQ3: Training Performance}
\label{sec:rq3_training}

We leverage \tool to generate solver-verified Chain-of-Thought (CoT) traces,
which encode explicit structural dependencies rather than stylistic reasoning patterns.
We fine-tune DeepSeek-R1-1.5B-Distill, GLM-4.1V-9B, and Qwen3-14B-Thinking using these verified traces.
Evaluation is conducted without access to formal tools at inference time.
For GSM8K and MATH, we report performance on both the original datasets and the corresponding \tool-generated benchmarks.
For Physics and Chemistry, since the original model settings do not include matched origin-dataset evaluations, we report results only on our \tool-generated benchmarks.
Table~\ref{tab:cot_training_effect} summarizes the results.
%%%%%%%%%
Overall, training with solver-verified CoT traces yields consistent improvements across models.
On the original GSM8K and MATH datasets, all three models improve after fine-tuning,
suggesting that verified structural supervision does not degrade standard benchmark performance.
% These results indicate that training on formally verified reasoning traces preserves, and in some cases strengthens,
% performance on the original natural-language benchmarks.
The gains are more pronounced on the \tool-generated benchmarks, where structural dependencies are explicitly controlled.
These results suggest that solver-verified CoT training helps models internalize structural reasoning dependencies,
rather than merely learning surface-level solution styles.
% Because formal tools are unavailable during inference, the observed improvements must come from parameters updated during training.
% Moreover, the consistent gains on both original benchmarks and \tool-generated benchmarks indicate that verified structural supervision can improve general reasoning robustness while preserving standard task performance.
The model-specific improvement patterns further suggest that different models benefit from verified CoT in different ways:
smaller models obtain stable but limited gains, whereas larger or mid-size reasoning models can exhibit substantial improvements when the training traces align with their latent structural reasoning capacity.
%%%%%%%%%%%
An interesting observation is that the strongest performance on the original datasets does not necessarily translate to stronger performance on the \tool-generated benchmarks.
DeepSeek-R1-1.5B-Distill achieves the highest accuracy among the three models on the original GSM8K and MATH datasets.
% reaching 83.0\% and 66.5\% after fine-tuning, respectively.
However, its performance drops substantially on the corresponding \tool benchmarks.
% with only 53.6\% on \tool-GSM8K and 31.1\% on \tool-MATH after fine-tuning.
This gap suggests that the 1.5B distilled model may be highly specialized to the distribution of original datasets,
possibly through distillation-induced memorization of common templates or dataset-specific reasoning patterns.
In contrast, larger models show lower performance on some original datasets
but stronger adaptation on structurally transformed \tool benchmarks.
This indicates that original benchmark accuracy and structured generalization capacity are not equivalent:
a model can perform well on familiar benchmark distributions while remaining brittle under controlled structural perturbations.

\begin{table*}[htbp]
\centering
\small
\caption{Effect of training with formally verified CoT on original datasets and \tool benchmarks (\% success rate).}
\label{tab:cot_training_effect}
% \resizebox{\textwidth}{!}{%
\begin{tabular}{l l|cc|cc|cc|cc}
\toprule
\multirow{2}{*}{\textbf{Model}} 
& \multirow{2}{*}{\textbf{Training}}
& \multicolumn{2}{c|}{\textbf{GSM8K}}
& \multicolumn{2}{c|}{\textbf{MATH}}
& \multicolumn{2}{c|}{\textbf{Physics}}
& \multicolumn{2}{c}{\textbf{Chemistry}} \\
\cmidrule(lr){3-4}
\cmidrule(lr){5-6}
\cmidrule(lr){7-8}
\cmidrule(lr){9-10}
& 
& \textbf{Origin} & \textbf{\tool}
& \textbf{Origin} & \textbf{\tool}
& \textbf{Origin} & \textbf{\tool}
& \textbf{Origin} & \textbf{\tool} \\
\midrule

\multirow{2}{*}{DeepSeek-R1-1.5B-Distill}
& Before 
& 80.0 & 49.3
& 62.5 & 28.8
& -- & 29.0
& -- & 24.6 \\
& After
& 83.0 & 53.6
& 66.5 & 31.1
& -- & 30.2
& -- & 24.7 \\
\midrule

\multirow{2}{*}{GLM-4.1V-9B}
& Before
& 61.0 & 43.0
& 44.5 & 33.9
& -- & 37.2
& -- & 39.0 \\
& After
& 63.5 & 77.0
& 47.5 & 35.5
& -- & 43.2
& -- & 44.8 \\
\midrule

\multirow{2}{*}{Qwen3-14B-Thinking}
& Before
& 71.0 & 64.8
& 56.5 & 40.6
& -- & 41.4
& -- & 28.2 \\
& After
& 75.0 & 68.5
& 58.5 & 42.5
& -- & 43.2
& -- & 58.3 \\

\bottomrule
\end{tabular}
% }
\end{table*}

%%%%%%%%%%%%%%%%%%%%%%%%%%%%%%%%%%%%%%%%%%%%%%%%%%%%%%%%%%%%%%%%%%%%%%%%%%%%%%%%%%%%%%%%%%%%%%%%%%%%%%%%%%%%%%%
\subsection{RQ4: Probes' Validity and Reliability}
% (Tri-verification success rate, failure modes)
This experiment assesses the reliability with which \tool generates valid, well-posed benchmarks. 
For each domain, we conduct tri-verification to check whether they (i) compile into executable formal code, (ii) dynamically execute the formal code and verify with the ground truth, and cross-check across different formal methods if they exist, and (iii) employ a judge LLM to confirm the semantics. 
In addition, we manually inspect 100 examples from each domain in the final dataset to further assess semantic correctness.
Results in Table~\ref{tab:rq3_success} report the success rate as the fraction of problems that pass each verification stage. 
Overall, the tri-verification experiment shows that high-quality problems are retained, maintaining both syntactic validity and semantic faithfulness, while the manual check further confirms the reliability of the generated probes.

\begin{table}[htbp]
\centering
\small
\caption{Success rate (\%) of automatic formal benchmark generation after tri-verification and manual checking.}
\label{tab:rq3_success}
\begin{tabular}{lcccc}
\toprule
\textbf{Domain} & \textbf{Static} & \textbf{Dynamic} & \textbf{Semantic} & \textbf{Human} \\
\midrule
GSM8K      & 97.62 & 91.42 & 96.29 & 98.00 \\
MATH       & 85.01 & 89.78 & 91.30 & 98.00 \\
Physics    & 91.23 & 83.17 & 97.30 & 98.00 \\
Chemistry  & 84.47 & 81.54 & 95.97 & 96.00 \\
\bottomrule
\end{tabular}
\end{table}
 % 3
\section{Discussion}
\label{sec:discussion}
% \paragraph{Error analysis.}
To better understand the nature of model failures, we conducted an error analysis with \tool on 100 incorrect responses each from o4-mini and GPT-4o across four datasets. 
For o4-mini, the results reveal a diverse taxonomy of failures.
% , evenly split between arithmetic breakdowns and logical collapses. 
Specifically, 34\% of the errors were classified as \textit{Numerical Calculation Errors}, where the model formulated the correct logic but failed in basic arithmetic execution. 
Another 34\% were due to \textit{Reasoning Chain Disruption}; for instance, the model would correctly identify the oxidation half-reaction but fail to carry through the multi-step derivation to combine with oxygen and balance the full redox equation, resulting in a critical logical gap. 
Furthermore, 27\% of failures were attributed to \textit{Logical Hallucination}, where the model fabricated constraints not present in the prompt (e.g., inventing an extra, unspecified third draw in a probability setup), leading to a completely invalid derivation. 
% The remaining 5\% fell into other minor categories. 
In stark contrast, GPT-4o's error profile reveals a severe vulnerability in sustaining deep derivations: its \textit{Reasoning Chain Disruption} surges to 51\%. 
Its remaining failures are distributed among \textit{Logical Hallucination} (18\%), \textit{Numerical Calculation Errors} (17\%), a  \textit{Convergence/Search Failure} mode (10\%) where the model fails to reach a conclusion in complex spaces, and other minor categories (4\%). 
This comparative breakdown highlights that as structural complexity increases, models do not merely guess randomly; their reasoning chains actively fracture or hallucinate, reinforcing the value of our explicit, solver-verified probe space for diagnosing vulnerabilities.

\paragraph{Limitations.}
Our study has several limitations.
First, while we verify probes to control for surface-level confounders, formalization inevitably abstracts away aspects of natural language that may matter in real-world reasoning tasks.
Second, the structural dimensions we study, such as constraint composition or dependency depth, are not exhaustive; other forms of complexity may induce different reasoning behaviors.
Finally, reasoning capabilities may depend on prompting strategies or interaction protocols, which we only partially explore.
Addressing these limitations will require extending probe families, integrating richer task formalisms, and closing the loop between measuring reasoning capabilities and training reasoning models.
More discussion is shown in \autoref{appendix}. % 0.5
\section{Related Work}\label{relaedwork}

\paragraph{Probing LLM Capability}
Large language models have demonstrated remarkable proficiency in solving complex academic and STEM-related problems, achieving human-level performance on rigorous university-level examinations~\cite{Drori2023FromHuman}. 
Despite these successes, a growing body of work investigates the underlying capabilities of these models by probing reasoning beyond aggregate accuracy metrics~\cite{ZHAO2025beyondaccuracy}. 
Dziri et al.~\citep{dziri2023faith} study the limits of transformer models for compositional reasoning, model multi-step reasoning tasks as computation graphs, and show that performance can degrade sharply as compositional complexity increases. 
Their results suggest that transformer models often rely on surface-level pattern matching rather than systematic composition, leading to brittle generalization under increased structural demands. 
Complementary to analyses based on task complexity, Bai et al.~\citep{bai2025} examine how reasoning behaviors vary across training stages and capability dimensions. 
They show that different post-training strategies induce qualitatively different generalization behaviors: while some fine-tuning methods preserve structured reasoning abilities, others increasingly cause models to rely on surface statistical patterns. 
A recent extensive data-driven survey~\cite{kostikova2025lllmsdatadrivensurveyevolving} confirms that reasoning failures remain the most prominent limitation of LLMs. 
Understanding these failures requires exploring the ``knowledge boundaries'' of LLMs, as models often exhibit limitations in memorizing and utilizing knowledge, which in turn lead to untruthful or inaccurate responses~\cite{li2025knowledgeboundarylargelanguage}. 
To systematically probe logical reasoning independent of domain knowledge, holistic benchmarks like LogiEval~\cite{liu2025evaluatinglogicalreasoningabilities} have been introduced, revealing that fundamental reasoning bottlenecks persist regardless of model scale. 
Furthermore, targeted evaluation frameworks such as LogicAsker~\cite{wan2024logicasker} adopt propositional and predicate logic to systematically expose vulnerabilities, demonstrating that LLMs struggle with specific logic rules and fail to perform consistent logical derivations, often memorizing patterns rather than genuinely learning the underlying rules. 
To address these limitations, the field is increasingly exploring the use of formal languages and verification methodologies to guide and assess LLM reasoning. 
Our work further advances this line of inquiry by explicitly operationalizing reasoning capacity as a function of extractable task structure, rather than treating compositionality or training strategy as implicit factors. 
% Unlike prior studies that analyze performance collapse on fixed or loosely parameterized tasks, we construct a formally grounded probe space in which structural properties are explicitly defined, calibrated, and solver-verified.

\paragraph{Benchmark and Evaluation for LLM}
Traditional benchmarks such as MATH~\citep{hendrycks2021math} and GSM8K~\citep{cobbe2021gsm8k} have been widely used to evaluate mathematical reasoning. 
However, as model accuracy approaches saturation (e.g., 87.9\% on MATH~\citep{lei2024macm} and 97.1\% on GSM8K~\citep{zhong2024gsm8k}), their discriminative power diminishes. 
Recent benchmarks, including ARB~\citep{sawada2023arb}, OlympiadBench~\citep{he2024olympiad}, and SciBench~\citep{wang2024scibench}, increase task difficulty but often rely on human evaluation. 
Efforts to automate grading with rubric-guided LLMs remain unreliable, motivating benchmarks such as Putnam-AXIOM~\citep{putnamaxiom2024} that provide standardized answers, enabling fully automatic evaluation.
On the other hand, contamination, in which evaluation data appear in pretraining corpora, has been widely recognized as a critical issue in LLM evaluation \citep{brown2020language,carlini2023quantifying,kandpal2022deduplicating}. 
MPA~\citep{dynamicevaluation} reformulates benchmark questions via paraphrasing and distractor insertion, with correctness verified by judge agents and human annotators. 
DYVAL~\citep{dyval} generates reasoning tasks using Directed Acyclic Graph structures, controlling difficulty through graph depth and width, while GSM-Symbolic~\citep{gsmsymbolic} expands GSM8K into symbolic templates with systematic surface variations. 
NPHardEval~\citep{nphardeval} constructs algorithmically verifiable problems across multiple complexity classes, and LiveBench~\citep{livebench} emphasizes contamination resistance by continuously refreshing tasks sourced from newly released materials.

Our work differs from these efforts by constructing benchmarks from formally specified probe programs whose latent structure is explicitly parameterized and solver-verified.
As a result, performance variation can be attributed to limits in structure extraction rather than data leakage, annotation noise, or evaluator inconsistency.

\paragraph{LLM Reasoning}
Recent work has shown that prompting strategies can substantially influence the reasoning behavior of large language models. 
Chain-of-Thought (CoT) prompting~\citep{wei2023chain} encourages models to generate intermediate reasoning steps before producing a final answer, leading to significant improvements on multi-step reasoning tasks. 
Wang et al.~\citep{wang2023self} propose self-consistency, which samples multiple reasoning paths under CoT prompting and aggregates their final answers. 
Beyond linear reasoning traces, Tree-of-Thoughts (ToT)~\citep{yao2023tree} generalizes CoT by explicitly modeling reasoning as a search process over a tree of intermediate states. 
By enabling branching, evaluation, and backtracking over partial solutions, ToT frames reasoning as structured exploration rather than a single left-to-right generation, providing a more flexible abstraction for complex problem solving.
Our work further complements these prompting-based approaches by focusing on what is being reasoned about, rather than how reasoning is elicited.
% \tool distinguish gains from improved search or variance reduction from genuine increases in reasoning capacity.

% \begin{table*}[t] % 双栏，强制在本页顶部
% \centering

% \caption{Comparison of recent related evaluation methods}
% \label{table:datasets}
% \small
% \begin{tabular}{l p{3.5cm} p{3.5cm} l} % p{width} 控制文字换行
% \toprule
% Title & Mutation & Validation & Templates Amount \\
% \midrule
% MPA & LLM/Human & LLM/Human & about 15000 \\
% \hline
% DYVAL & Predefined rules  & DAG Algorithm & - \\
% \hline
% GSM-Symbolic & Predefined rules & Human and Algorithm & 50 \\
% \hline
% NPHardEval & Monthly update & Algorithm & 900 \\
% \hline
% LIVEBENCH & Monthly update & Automated scoring tools & about 1000 \\
% \hline
% \tool(ours) & Difficulty-aware & Formal Verification & about 9000 \\
% \bottomrule
% \end{tabular}
% \end{table*}

 % 1
\section{Conclusion}
\label{sec:conclusion}
This paper introduced \tool, a measurement-oriented framework for quantifying the structural reasoning capabilities of LLMs using formalized and calibrated probes.
% Our results demonstrate that structurally controlled evaluation reveals distinctions between models that remain invisible under standard aggregate metrics. 
% By systematically varying compositional depth, constraint interaction, and solution-space organization, \tool enables precise localization of brittle reasoning operations and structurally interpretable failure modes.
Looking forward, \tool opens several directions for future work. 
On the modeling side, calibrated structural measurements can guide the design of training and the targeted improvement of fragile reasoning components. 
On the evaluation side, formally verified probes provide a formalized foundation for testing reasoning LLMs in safety-critical settings.
% More broadly, we advocate for a shift from surface-level accuracy benchmarks toward evaluation protocols grounded in explicit structural variation and formal verification. 
% Such an approach offers a more reliable lens for studying generalization, robustness, and the evolving reasoning behavior of large language models. % 

\begin{acks}
This study was supported by the Ministry of Education,
Singapore under its Academic Research Fund Tier 3 (MOET32020-0004) and the Cyber Security Agency under its National Cybersecurity R\&D Programme (NCRP25-P04-TAICeN).
\end{acks}

\newpage
\bibliographystyle{ACM-Reference-Format}
\bibliography{reference}

\appendix

\section{More Discussion}
\label{appendix}

\paragraph{Beyond correctness.}
Formalization contributes more than answer verification. By making task structure explicit, it provides a common reference frame for comparing models, prompts, and failure modes. It turns vague notions of reasoning difficulty into analyzable structural factors such as constraint composition, dependency depth, and cross-step coupling. As a result, failures near a capability frontier can be traced to specific structural requirements rather than being attributed only to noise or prompt sensitivity.

\paragraph{Stepwise reasoning progress.}
The frontier patterns observed in our experiments suggest that reasoning ability often improves in discrete structural regimes rather than through smooth degradation or continuous gains. Models may remain stable under moderate structural growth, but fail abruptly once a required extraction or recomposition pattern exceeds their effective capacity. This motivates staged evaluation and training, where calibrated probes act as milestones for identifying which structures have been internalized and which remain brittle.

\paragraph{Frontiers as behavioral maps.}
Capability frontiers should therefore be viewed not as single-number metrics, but as structured descriptions of model behavior. Aggregate accuracy compresses performance and can obscure important differences between models with similar scores. Frontier analysis instead asks which structures a model can reliably extract, preserve, and compose. In this sense, frontiers function as behavioral maps: they reveal where reasoning is stable, where it becomes brittle, and which structural axes drive the transition.

\paragraph{Implications for supervision.}
Although our focus is evaluation, the same framework naturally informs training. Solver-verified probes can provide high-precision supervision targeted at frontier-adjacent structures. This enables closed-loop paradigms in which capability measurements guide data generation, curriculum design, or fine-tuning, aiming to shift capability boundaries systematically rather than merely improve aggregate performance.

\begin{figure}[htbp]
    \centering
    \includegraphics[width=\columnwidth]{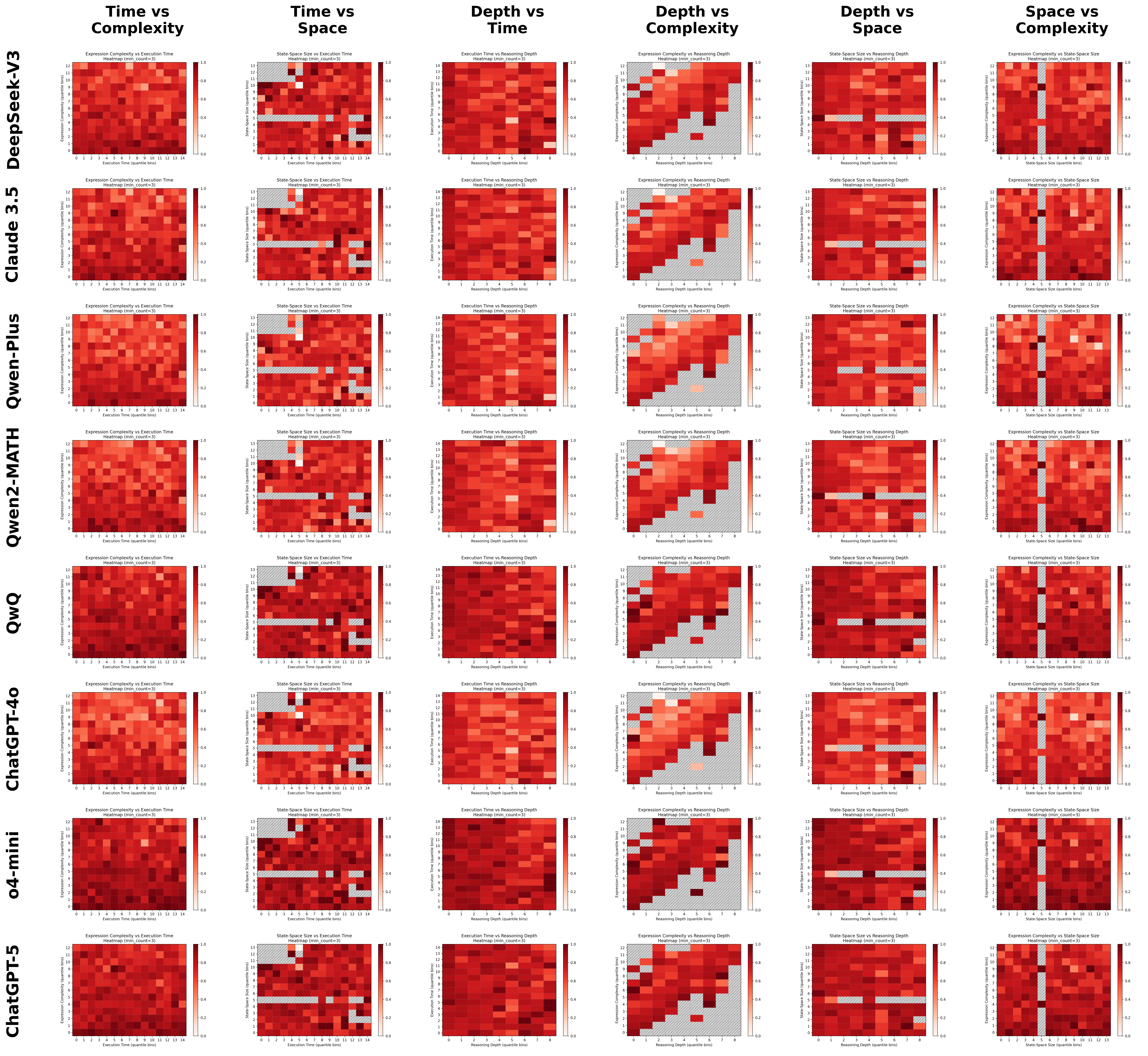}
    \caption{\textbf{GSM8K Dataset}: Heatmap visualization of pairwise structures in model success rates.}
    \label{fig:combined_gsm8k}
\end{figure}

\begin{figure}[htbp]
    \centering
    \includegraphics[width=\columnwidth]{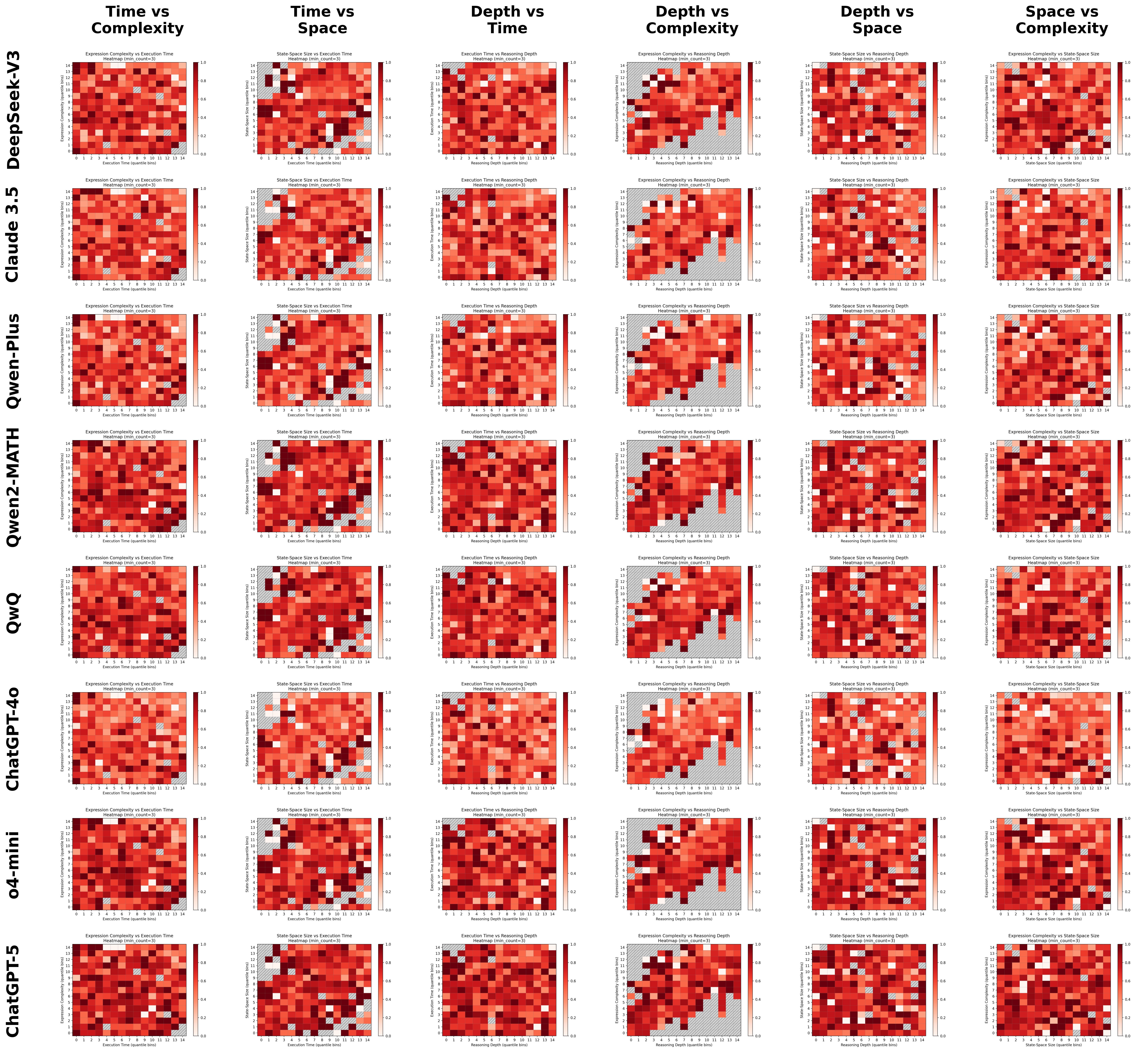}
    \caption{\textbf{MATH Dataset}: Heatmap visualization of pairwise structures in model success rates.}
    \label{fig:combined_math}
\end{figure}

\begin{figure}[htbp]
    \centering
    \includegraphics[width=\columnwidth]{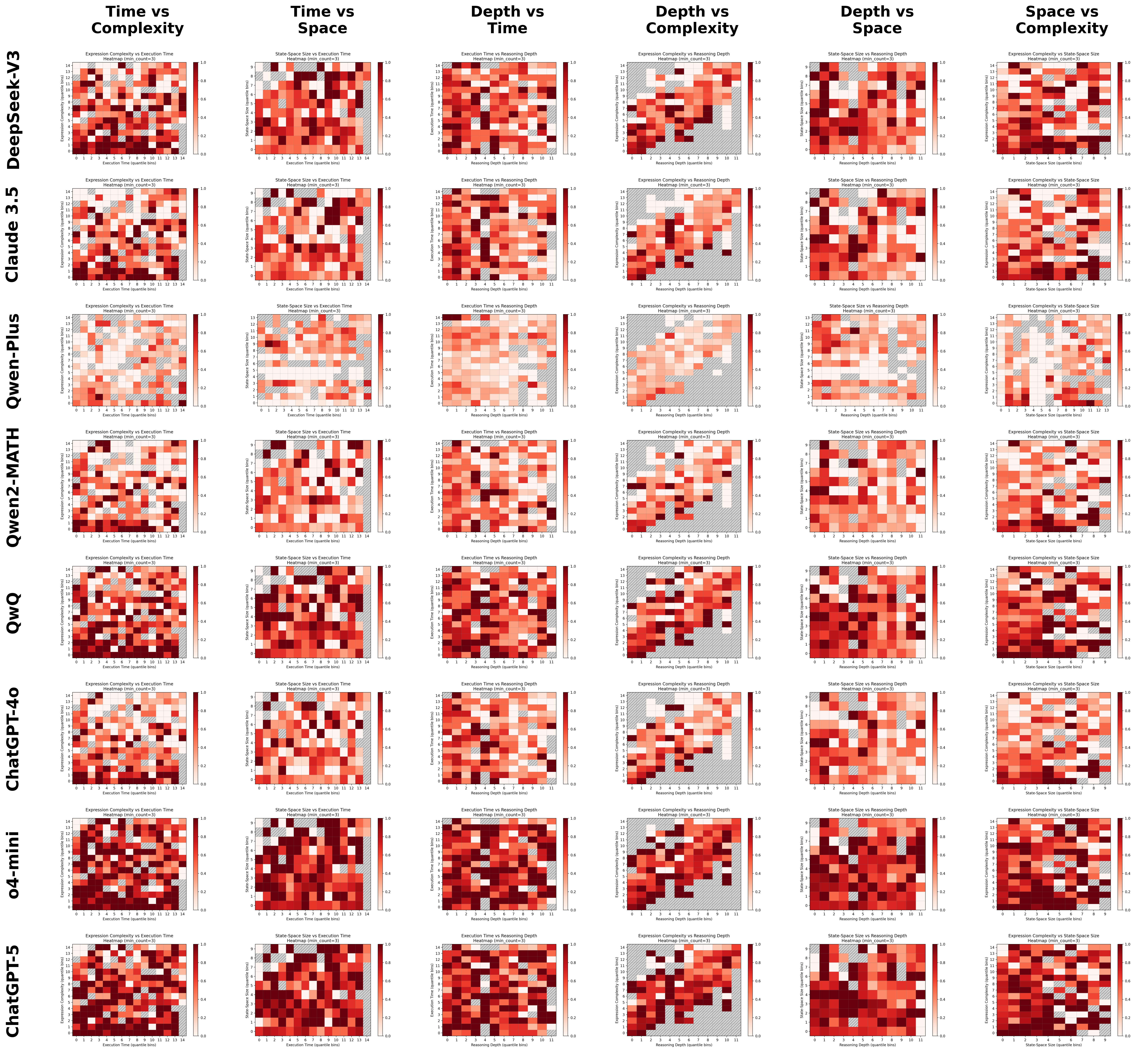}
    \caption{\textbf{Physics Dataset}: Heatmap visualization of pairwise structures in model success rates.}
    \label{fig:combined_physics}
\end{figure}

\begin{figure}[htbp]
    \centering
    \includegraphics[width=\columnwidth]{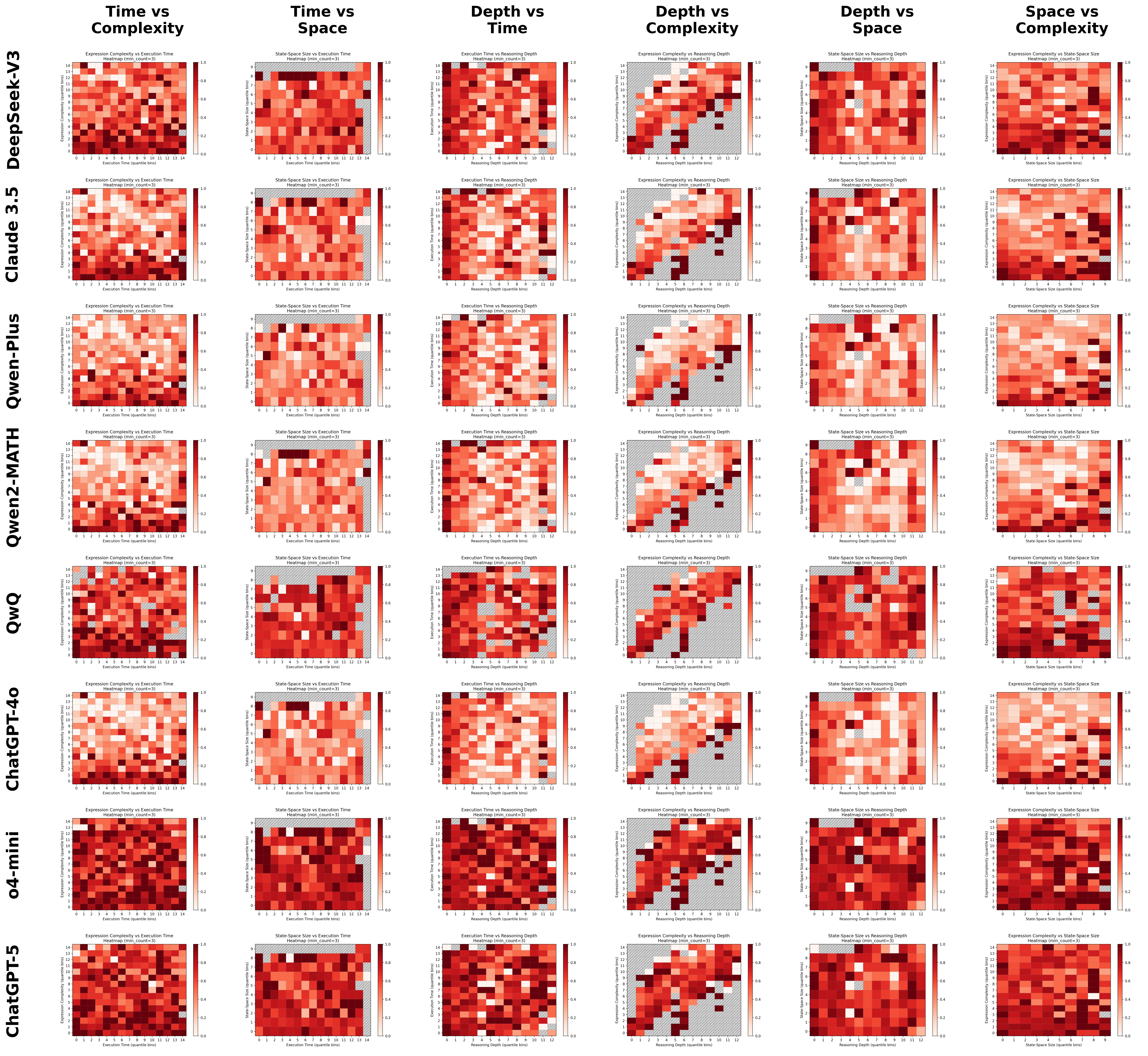}
    \caption{\textbf{Chemistry Dataset}: Heatmap visualization of pairwise structures in model success rates.}
    \label{fig:combined_chemistry}
\end{figure}

\begin{table}[htbp] 
\centering 
\small 
\caption{Average per-sample token cost and inference time across four datasets.} 
\label{app_token_usage} 
\resizebox{\columnwidth}{!}{%
\begin{tabular}{lcc|cc} 
\toprule 
\textbf{Model} & \multicolumn{2}{c|}{\textbf{GSM8K}} & \multicolumn{2}{c}{\textbf{MATH}} \\ 
\cmidrule(lr){2-3} \cmidrule(lr){4-5} 
& \textbf{Avg Tokens} & \textbf{Time} & \textbf{Avg Tokens} & \textbf{Time} \\ 
\midrule 
Claude-3.5 & 132.36 & 4.40s & 236.44 & \textbf{4.83s} \\ 
DeepSeek-V3 & 178.12 & 10.50s & 532.62 & 20.30s \\ 
GPT-4o & 203.83 & \textbf{3.48s} & 404.33 & 6.25s \\ 
o4-mini & \textbf{24.08} & 5.45s & 73.61 & 12.11s \\ 
GPT-5 & 50.62 & 26.6s & \textbf{70.56} & 32.61s \\ 
Qwen-Plus & 172.72 & 11.42s & 499.70 & 26.67s \\ 
Qwen2-Math & 173.21 & 7.32s & 525.41 & 13.21s \\ 
QwQ & 46.77 & 18.85s & 253.60 & 30.85s \\ 
\midrule 
\textbf{Model} & \multicolumn{2}{c|}{\textbf{PHYSICS}} & \multicolumn{2}{c}{\textbf{CHEMISTRY}} \\ 
\cmidrule(lr){2-3} \cmidrule(lr){4-5} 
& \textbf{Avg Tokens} & \textbf{Time} & \textbf{Avg Tokens} & \textbf{Time} \\ 
\midrule 
Claude-3.5 & 191.08 & \textbf{3.52s} & 168.23 & \textbf{4.13s} \\ 
DeepSeek-V3 & 325.21 & 12.55s & 269.36 & 13.10s \\ 
GPT-4o & 335.72 & 5.85s & 295.73 & 6.50s \\ 
o4-mini & 67.51 & 15.88s & \textbf{31.91} & 15.15s \\ 
GPT-5 & \textbf{58.87} & 23.07s & 46.14 & 33.67s \\ 
Qwen-Plus & 386.17 & 18.29s & 291.93 & 17.14s \\ 
Qwen2-Math & 423.39 & 10.88s & 317.33 & 15.30s \\ 
QwQ & 236.90 & 16.81s & 194.40 & 16.32s \\ 
\bottomrule 
\end{tabular} 
}
\end{table}

\section{Computational Cost}\label{app:cost}

\autoref{app_token_usage} reports the average per-sample token consumption and inference latency across four domains.
Clear efficiency trade-offs emerge across models.
o4-mini achieves the lowest token usage on GSM8K and CHEMISTRY, requiring only 24.08 and 31.91 tokens per sample on average, respectively.
This indicates a compact reasoning trace with minimal verbosity.
On PHYSICS and MATH, GPT-5 exhibits the lowest average token consumption (58.87 and 70.56 tokens), suggesting that its reasoning traces remain relatively concise even on more complex domains.
In contrast, GPT-4o and Claude-3.5 achieve the lowest inference latency.
GPT-4o reaches the fastest runtime on GSM8K (3.48s), while Claude-3.5 achieves the lowest latency on PHYSICS (3.52s) and CHEMISTRY (4.13s).
This indicates that lower token usage does not necessarily translate into lower wall-clock inference time.
Models such as DeepSeek-V3 and Qwen-Plus incur substantially higher token costs, particularly on MATH, where the average token usage exceeds 500 tokens per sample.
Similarly, GPT-5 maintains moderate token usage but comparatively higher latency, indicating heavier computational overhead during inference.
Overall, token efficiency and time efficiency are not perfectly aligned.
Models differ in verbosity, decoding strategies, and internal reasoning depth, resulting in distinct cost–performance trade-offs across structured domains.

\section{Online Resources}
We publish our framework, benchmark, and supplementary materials on the project website~\citep{ace-nips}.
The website serves as an interactive extension of the paper. 
It provides a broader collection of generated probes, including additional examples, visual illustrations, videos, and boundary cases that are difficult to fully include in the main text.
The website also includes runnable artifacts and benchmark files to facilitate reproduction and future comparison.

\end{document}